# Mask Editor : an Image Annotation Tool for Image Segmentation Tasks

Chuanhai Zhang, Kurt Loken, Zhiyu Chen, Zhiyong Xiao, Gary Kunkel

{chuanhai.zhang, kurtis.d.loken, zhiyu.chen, zhiyong.x.xiao, gary.j.kunkel}@seagate.com

Seagate company

*Abstract*— **Deep convolutional neural network (DCNN) is the state-of-the-art method for image segmentation, which is one of key challenging computer vision tasks. However, DCNN requires a lot of training images with corresponding image masks to get a good segmentation result. Image annotation software which is easy to use and allows fast image mask generation is in great demand. To the best of our knowledge, all existing image annotation software support only drawing bounding polygons, bounding boxes, or bounding ellipses to mark target objects. These existing software are inefficient when targeting objects that have irregular shapes (e.g., defects in fabric images or tire images). In this paper we design an easy-to-use image annotation software called Mask Editor for image mask generation. Mask Editor allows drawing any bounding curve to mark objects and improves efficiency to mark objects with irregular shapes. Mask Editor also supports drawing bounding polygons, drawing bounding boxes, drawing bounding ellipses, painting, erasing, super-pixel-marking, image cropping, multi-class masks, mask loading, and mask modifying.**

*Keywords—Image segmentation; defect detection; multi-class mask, deep learning; neural network*

## I. INTRODUCTION

Image segmentation is one of the key challenging computer vision tasks in computer science. Deep convolutional neural network (DCNN) [1, 2] is the state of the art method for image segmentation. However, DCNN requires a lot of training images with corresponding image masks to get a good segmentation result. Therefore, there is a need for an easy-to-use image annotation software to help fast image mask generation. On one hand, it is often that the object to be marked has irregular shapes (e.g., defects in fabric images or tire images, or hard disk images). On the other hand, there usually exist a lot of small defects in one image, which makes the manual defect marking process time-consuming. A lot of image annotation software exist [3-12] to help generate image masks. However, all these software only allow drawing bounding box, bounding ellipse, or bounding polygon to select and mark an object. These software are very limited and inefficient when dealing with objects with irregular shapes. For example, bounding polygon is the best existing software to mark an object with irregular shape. However, we have to carefully select many polygon vertices along the boundary of the irregular-shaped object to get an accurate mask. Additionally, we often need to load and rectify image masks. For example, in incremental learning or active learning, we need to rectify the predicted image masks of low quality and add them into the training set to improve the segmentation accuracy. Unfortunately, all existing image annotation software reviewed do not support image mask loading and editing.

In this paper, we design an easy-to-use image annotation software called Mask Editor for image mask generation. Mask Editor supports many features as follows.

- Freehand drawing, which allows marking a target object of *any irregular shape* by drawing only a bounding curve along the object's boundary

- Mark objects by drawing bounding boxes, bounding ellipses, bounding polygons

- Mark objects by directly painting on them

- Erasing wrong marking to get an accurate object mask

- Super-pixel-marking, which allows fast marking small irregular objects by single click on them

- Multi-class masks

- Load and edit existed image masks

- Crop image to extract regions of interest (ROI)

- Zoom in, zoom out, and pan

- Save masks

- Navigate to the previous or next image

## II. RELATED WORK

Currently, there exist a lot of image annotation software [3-12] to help generate image masks. All these software mark objects by drawing bounding boxes, bounding ellipses, or bounding polygons. In Table I, we list important features (the first column) that an ideal image annotation software is expected to support, then summarize and compare different annotation software. The feature list is explained as follows.

Table I. Summary on all existing image annotation software

| Software / Features | Mask Editor [13] | Dataturks [3] | OCLAVI [4] | LabelBox [5] | LabelImg [6] | Pixorize [7] | LableMe [8] | Annotorious [9] | Ratsnake [10] | VIA [11] | ImageTagger [12] |
|---|---|---|---|---|---|---|---|---|---|---|---|
| *B-box* | ✓ | ✓ | ✓ | ✓ | ✓ | ✓ | ✗ | ✓ | ✗ | ✓ | ✓ |
| *B-polygon* | ✓ | ✓ | ✓ | ✓ | ✗ | ✓ | ✓ | ✓ | ? | ✓ | ✓ |
| *B-ellipse* | ✓ | ✗ | ✓ | ✗ | ✗ | ✓ | ✗ | ✗ | ✗ | ✓ | ✗ |
| *B-curve* | ✓ | ✗ | ✗ | ✗ | ✗ | ✗ | ✗ | ✗ | ✗ | ✗ | ✗ |
| *Painting* | ✓ | ✗ | ✗ | ✗ | ✗ | ✗ | ✗ | ✗ | ✓ | ✗ | ✗ |
| *Erasing* | ✓ | ✗ | ✗ | ✗ | ✗ | ✗ | ✗ | ✗ | ✗ | ✗ | ✗ |
| *Delete-mark* | ✓ | ✓ | ✓ | ✓ | ✓ | ✓ | ✓ | ✓ | ? | ✓ | ✓ |
| *Super-pixel-mark* | ✓ | ✗ | ✗ | ✗ | ✗ | ✗ | ✗ | ✗ | ✗ | ✗ | ✗ |
| *Cropping* | ✓ | ✗ | ✗ | ✗ | ✗ | ✗ | ✗ | ✗ | ✗ | ✗ | ✗ |
| *Multi-class* | ✓ | ✓ | ✓ | ✓ | ✓ | ✗ | ✓ | ✓ | ? | ✓ | ✓ |
| *Zoom in/out* | ✓ | ✗ | ✓ | ✓ | ✓ | ✓ | ✓ | ✓ | ✓ | ✓ | ? |
| *Pan* | ✓ | ✗ | ✓ | ✓ | ✓ | ✓ | ✓ | ✓ | ✓ | ✓ | ? |
| *Load/Edit-mask* | ✓ | ✗ | ✗ | ✗ | ✗ | ? | ? | ? | ? | ? | ? |
| *Save-mask* | ✓ | ✓ | ✓ | ✓ | ✓ | ? | ✓ | ? | ? | ✓ | ✓ |
| *License* | Free | Custom | Custom | Custom | MIT | Custom | GPL | MIT | Custom | BSD | MIT |

Notes: ✓ represents the software **supports** the feature; ✗ represents the software **does not support** the feature; ? represents **not sure**.

**B-box, B-polygon, and B-ellipse** represent marking an object by drawing a bounding box, a bounding polygon, and a bounding ellipse, respectively. They are commonly used marking operations.

**B-curve** represents marking an object by drawing a bounding curve along the object's boundary. The bounding curve may have any shape. B-curve is very useful to mark objects with complex and irregular shapes.

**Painting** represents brushing directly on an image to mark the target object. **Erasing** represents brushing directly on an image to remove the wrong markings. Painting is useful when the target object has thin and irregular shapes. Both painting and erasing are expected to support changing the size of the brush. Painting and erasing are quite helpful to get the accurate mask of an object, especially on the object boundary.

**Delete-mark** represents undo the last object marking, it is very necessary since we may make mistakes when marking objects.

**Super-pixel-marking** represents marking a small region by single click on it. Super-pixel-marking is very helpful when marking small and irregular objects.

**Cropping** represents extracting only regions of interest (ROI) of the loaded image rather than the whole image for editing, cropping is very useful when the ROI only occupied a small part of the whole image.

**Multi-class** represents supporting marking objects from different object classes.

**Zoom in, zoom out, and pan** allow us to enlarge and move the target object to the center of working area so that we can accurately mark that object.

**Load/Edit-mask** represents loading and editing the existing image masks. In incremental learning or active learning, we may need to rectify the predicted image mask of low quality and add them into the training set to improve the segmentation accuracy

**Save-mask** represents saving the generated masks of the loaded image by the annotation software into a file.

### III. FEATURES OF MASK EDITOR

Mask Editor is developed using Matlab based on an open source program called Multi ROI Editor [14]. Mask Editor uses the Matlab image processing toolbox. Figure 1 shows the work-flow of Mask Editor. Figure 2 shows the main GUI of Mask Editor. In this section, we will describe key features of our Mask Editor in details.

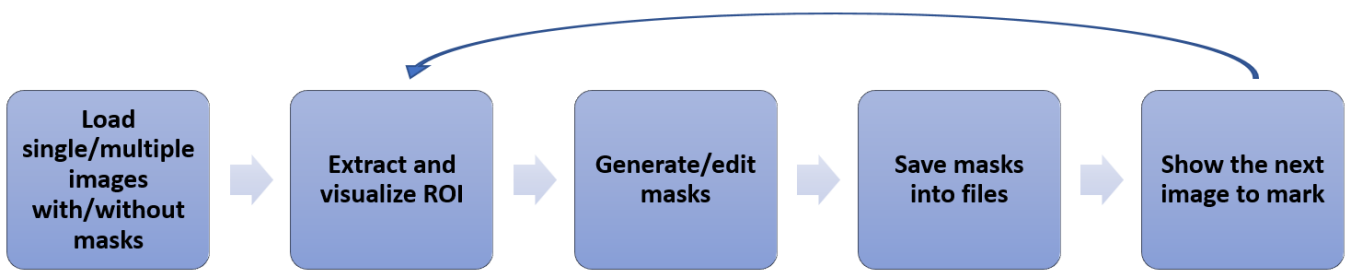

Fig.1. The work-flow of our Mask Editor

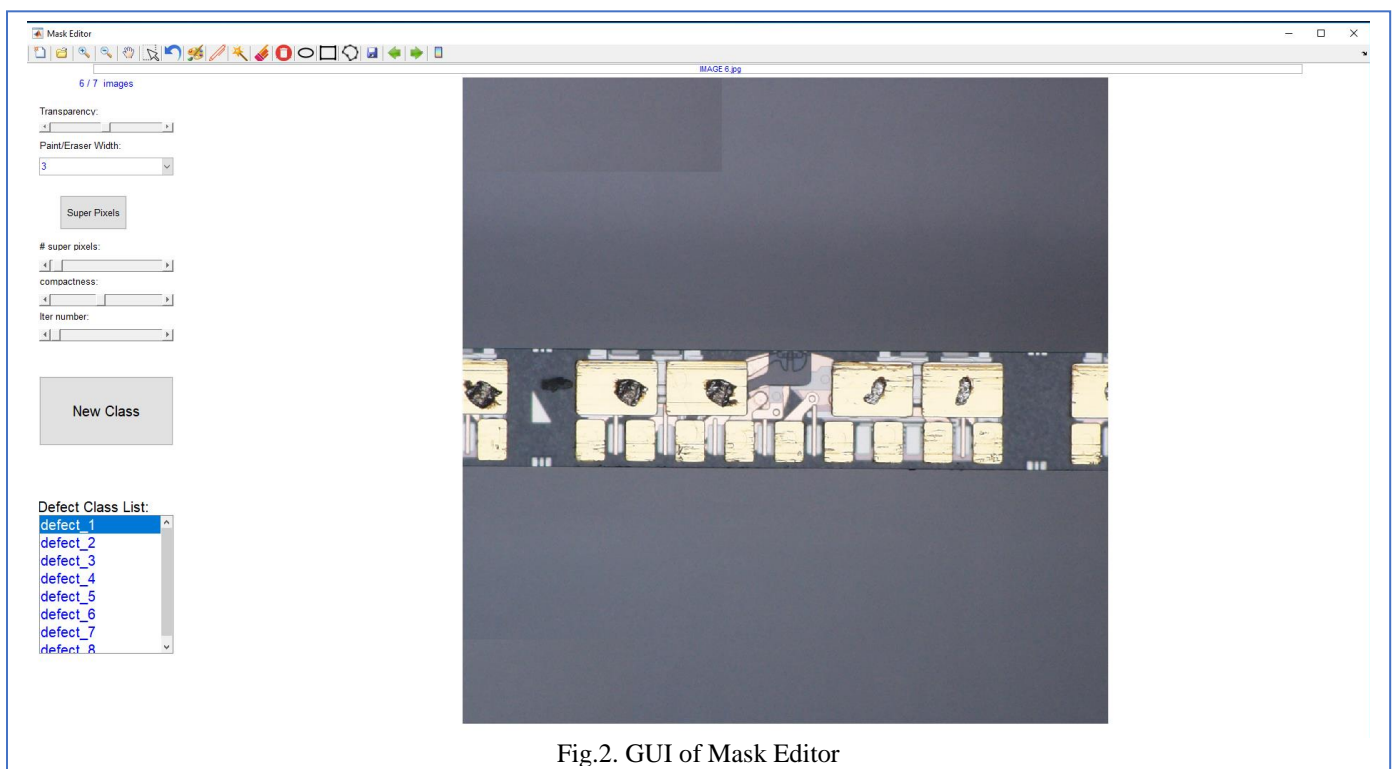

Fig.2. GUI of Mask Editor

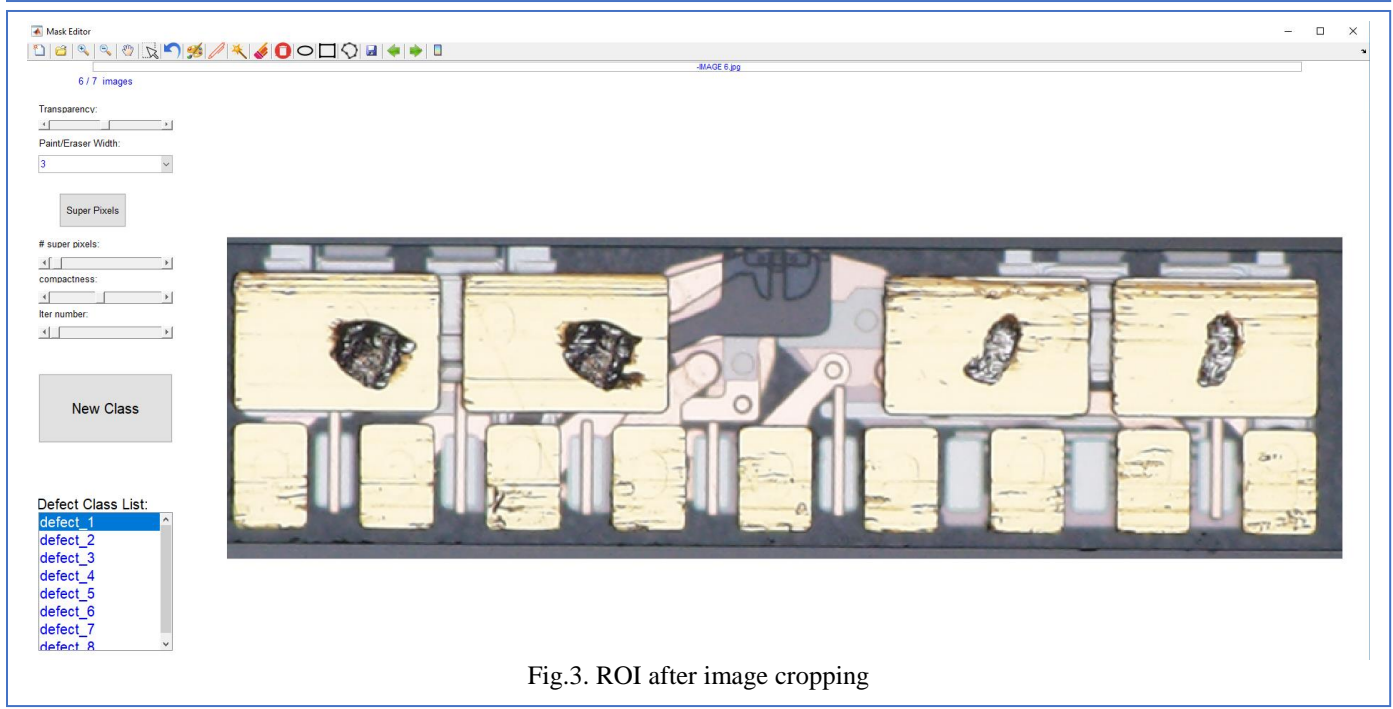

Fig.3. ROI after image cropping

*A. Image/Mask Loading*

Mask Editor supports loading four types of images (JPG, PNG, BMP, and TIFF). Mask Editor supports loading multiple selected images of different formats at the same time. It also supports loading all images in a folder. Mask Editor automatically loads pre-defined masks of the current image.

*B. Image Cropping*

In many situations (e.g., marking defects in industrial images), target objects we want to mark only appear in a local area of the whole image. We call this local area the region of interest (ROI). Before marking objects, it is highly recommended to crop the image and only show the ROI in the working area as shown in Fig. 3. Cropping contributes a lot to real-time visualization of markings. Mask Editor allows cropping raw images multi times to get multiple ROIs. When finishing the current ROI marking, we just need to click to go back the raw image and select a new ROI for marking. Mask Editor will save markings on previous ROIs automatically.

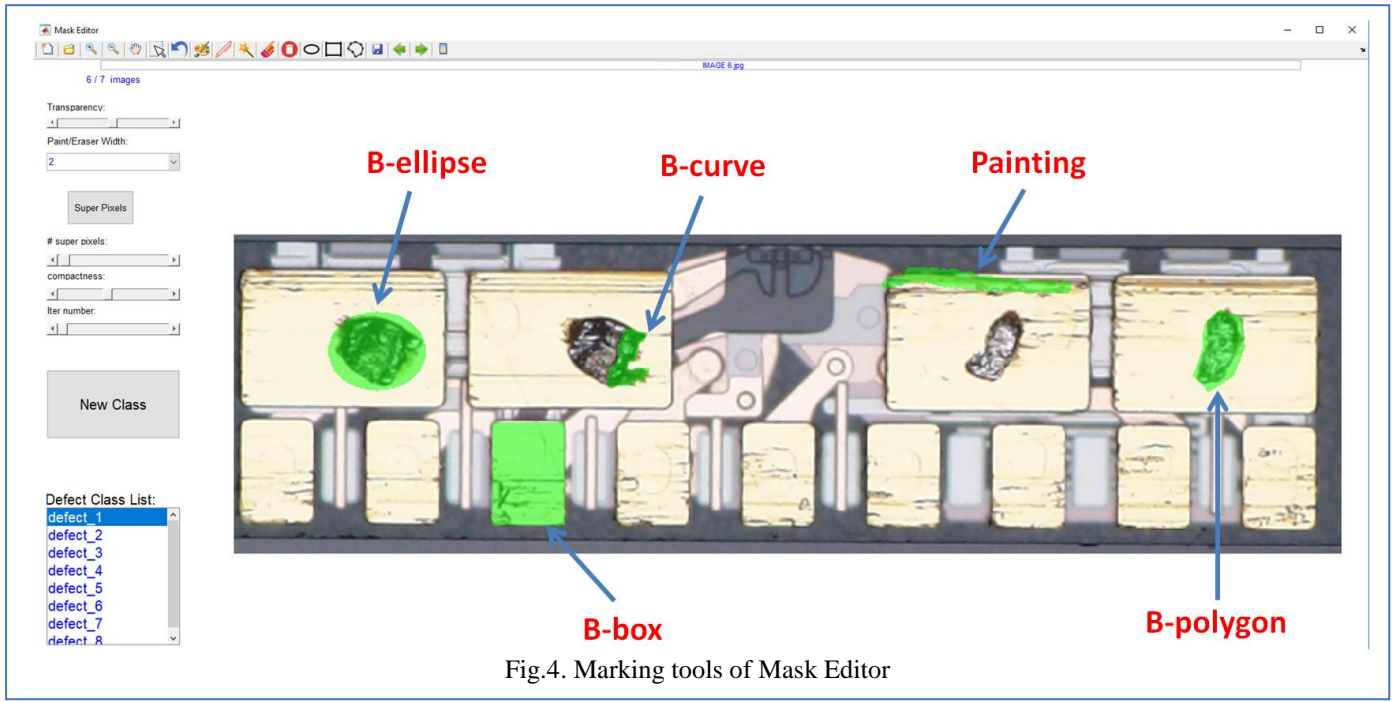

Fig.4. Marking tools of Mask Editor

*C. Mask Creation/Modificaiton*

For image segmentation tasks, each object to mark should have a class name. Before starting to mark an object, we need to name that object. Mask Editor automatically preloads a list of different object class names into a list box in GUI as shown in Fig. 1. If we find a new object class never seen before, we just need to create a new class name and add into the object class list. Then, we just need to select one class name and begin marking objects belonging to that class. Mask Editor visualizes all marked objects belonging to the current object class in real time. Mask Editor also supports marking-switch between different object classes by clicking certain object class name in the class list. Mask Editor supports changing the mask color and transparency as well.

*1) Mask Creation:* Mask Editor provides five types of operations to mark an object. They are B-box □, B-polygon ⬡, B-ellipse ○, B-curve ✎, Painting 🖌 as shown in Fig. 4, and Super-pixel-mark ✨ as shown in Fig. 5.

- **B-box** □ , **B-polygon** ⬡ , and **B-ellipse** ○ represent marking an object by drawing a bounding box, a bounding polygon, and a bounding ellipse, respectively. They are the main operations to mark objects used in existing image annotaiton softwares.

- **B-curve** ✎ represents marking an object by drawing a bounding curve along the object's boundary. The bounding curve may have any shape. B-curve is useful when the object to mark has an irregular shape.

- **Painting** 🖌 represents brushing directly on an image to mark the target object. Painting is useful when the target object has thin and irregular shapes. Painting supports changing the size of the brush.

- **Super-pixel-marking** ✨ represents marking a small region by a single click on it. Before marking super pixels, we need to divide the image into super pixels as shown in Fig. 5 by clicking the button . Mask Editor directly invokes the simple linear iterative clustering (SLIC) algorithm [15] implemented in Matlab. We can setup three parameters (number of super pixels, compactness, iteration number) of SLIC by dragging the sliders. Mask Editor provides real-time visulizaiton of super pixels when dragging the sliders. Super-pixel-mark is very helpful when marking small and irregular objects.

*2) Mask Modificaiton:* Mask Editor provides two types of operations to modify an existing mask. They are "Erasing" and "Delete-mark".

- **Erasing** ✎ represents brushing directly on an image to remove the wrong markings. Eraser supports changing the size of the brush. Eraser is helpful to get the accurate mask of an object, especially on the object boundary.

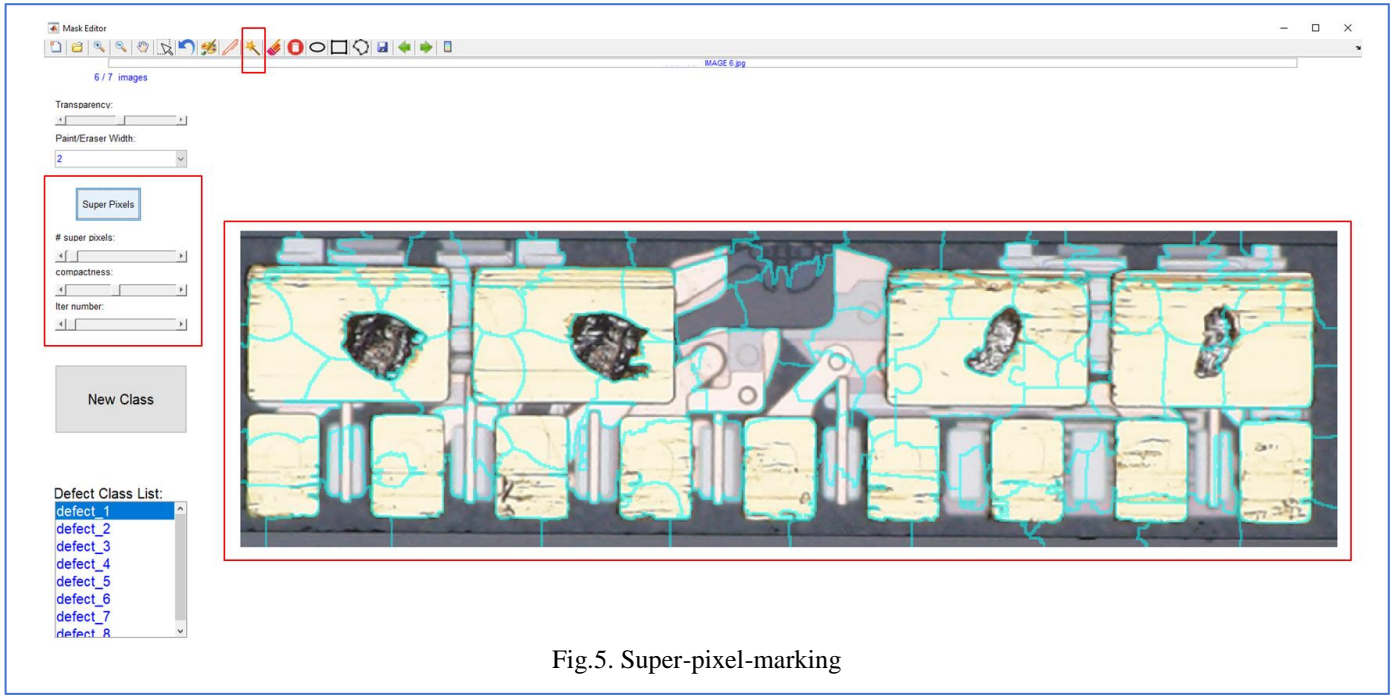

Fig.5. Super-pixel-marking

- **Delete-mark**  supports removing low-quality marks by drawing only a bounding-box. It is much more powerful than undo-marking operation, which only allows cancelling the previous marking operation.

*D. Saving Masks*

Mask Editor supports exporting generated masks of the loaded image to binary image files. Mask Editor saves one binary mask image for each object class. Mask Editor generates global masks corresponding to the original full image although it crops the original full image and edit only on ROI.

*E. Image nevigation*

Mask Editor supports automatically loading the previous or next image. We just need to click  or press the left key on the keyboard to show the previous image and click  or press the right key on the keyboard to show the next image. Mask Editor automatically saves the mask image files for the current image before loading another image. Mask Editor also automatically loads all the existed mask image files of the new image when navigating to the new image. It is very convenient when we finish marking the current image and want to load another image to mark.

*F. Program recovery from checkpoints*

Mask Editor automatically saves the number of images, which have been checked or marked, into a checkpoint file. If Mask Editor crashes for some unexpected reasons, we just need to restart the program. Then Mask Editor will automatically recover by reading the checkpoint files. Mask Editor will skip all previous checked or marked images and load images starting from the last image checked when the program crashed before.

## IV. CONCLUSION

In this paper, we design an image annotation software called Mask Editor. It is easy to use and allows fast image mask generation. Different from all existing image annotation software reviewed, Mask Editor provides three unique marking tools: B-curve, super-pixel-marking and erasing. These three marking tools are very helpful to mark objects with irregular shapes accurately and quickly. For the future work, we will explore more new features (e.g., build defect templates library and synthesize images by combining normal images and defect templates).